\documentclass{new_tlp}

\newcommand\bcmdtab{\noindent\bgroup\tabcolsep=0pt%
  \begin{tabular}{@{}p{10pc}@{}p{20pc}@{}}}
\newcommand\ecmdtab{\end{tabular}\egroup}

\usepackage{times}
\usepackage{helvet}
\usepackage{courier}
\usepackage{algorithm} 
\usepackage{algorithmic} 
\usepackage{microtype}
\usepackage{url}
\usepackage{latexsym}
\usepackage{graphicx}
\usepackage{graphics}
\usepackage{amsmath}
\usepackage{amssymb}
\usepackage{longtable}
\usepackage{multirow}
\usepackage{verbatim}
\usepackage{xspace}
\newtheorem{example}{Example}
\newtheorem{definition}{Definition}

\newtheorem{theorem}{Theorem}
\newtheorem{corollary}{Corollary}
\newcommand{\ignore}[1]{}

\newcommand{\ie}[1] {
  \begin{itemize}
    #1
  \end{itemize}
}

\newcommand{\ee}[1] {
  \begin{enumerate}
    #1
  \end{enumerate}
}

  \title[ESmodels: An Epistemic Specification Solver]
        {ESmodels: An Epistemic Specification Solver}

  \author[Z-Z. Zhang, K-K. Zhao]
         {ZHIZHENG ZHANG and KAIKAI ZHAO\\
         School of Computer Science and Engineering\\
         Southeast University, NanJing 211189, China\\
         \email{seu\_zzz@seu.edu.cn}}

\jdate{March 2003}
\pubyear{2003}
\pagerange{\pageref{firstpage}--\pageref{lastpage}}
\doi{S1471068401001193}

\begin{document}

\label{firstpage}

\maketitle

  \begin{abstract}
\centerline{\emph{(To appear in Theory and Practice of Logic Programming (TPLP))}}

$ESmodels$ is designed and implemented as an experiment platform to investigate the semantics, language, related reasoning algorithms, and possible applications of epistemic specifications. We first give the epistemic specification language of $ESmodels$ and its semantics. The language employs only one modal operator K but we prove that it is able to represent luxuriant modal operators by presenting transformation rules. Then, we describe basic algorithms and optimization approaches used in $ESmodels$. After that, we discuss possible applications of $ESmodels$ in conformant planning and constraint satisfaction. Finally, we conclude with perspectives.
  \end{abstract}


\section{Introduction}
The language of epistemic specification initially proposed in \cite{GelfondP91}, \cite{gelfond1993reasoning}, \cite{gelfond1994logic}, and \cite{491768} is an extension of the language of answer set programs by modal operators $\text{K}$ and $\text{M}$ to represent beliefs of the agent capable of introspection in the presence of multiple belief sets. Intuitively, it use $\text{K}F$ to denote an proposition $F$ is believed to be true in each of the agent's belief sets, and $\text{M}F$ to denote an proposition $F$ is believed to be true in some of the agent's belief sets. This extension is believed to be useful by discussing its application to formalization of commonsense reasoning. Along its syntax and semantics in \cite{GelfondP91}, a few efforts were made to establish reasoning algorithms in \cite{zhang2006computational} and \cite{watson1994}, and theoretical foundation in \cite{zhang2003minimal}, \cite{watson2000splitting}, and \cite{wang2005nested}. Recently, research on epistemic specifications increases again because introspective reasoning is becoming reality and forseeable as showed in \cite{faber2011manifold}, \cite{faber2009manifold}, and \cite{truszczynski2011revisiting}. To eliminate some unintended interpretations which exist under the original definition, a new semantics is defined in \cite{gelfond2011new} to arguably close to the intuitive meaning of modalities. Currently, efforts are still desired to made to establish and validate properties of epistemic specifications and the corresponding reasoning algorithms, and to investigate the use of the language. The design and implementation of an epistemic specification solver is hoped to facilitate those efforts.

This article introduces an epistemic specification solver $ESmodels$ that is recently being designed and implemented as a flexible platform for experiment with epistemic specifications. The language of $ESmodels$ has two types of subjective literals K$l$ and $\neg$K$l$. To express other types of subjective literals, we propose a group of transformation rules rewriting epistemic specifications with arbitrary types of subjective literals in $ESmodels$'s language. In $ESmodels$, a generate-test algorithm for computing world views of the epistemic specification is employed. It is worth noting that efficient ASP solver $Clasp$ is coupled into $ESmodels$ to help to generate candidate world views efficiently. Optimization approaches are preliminarily used to promoting the efficiency of the basic algorithm. Presently, we are applying $ESmodels$ in solving security conditions in conformant planning, and encoding constraint satisfaction problems.

\section{Language}
\subsection{Syntax and Semantics}
An $ESmodels$'s epistemic specification is a collection of finite rules in the following form
\[l_0\; or...or\; l_k :-\; l_{k+1},...,l_{j},~\text{S}l_{j+1},...,~\text{S}l_{m},\;not\;l_{m+1},...,\;not\;l_{n}\]
where each $l_{i}$ for $0\le i\le n$ is an \emph{objective literal}, ie. either an atom $A$ or the negation $\neg A$ of $A$, and $\text{S}$ is either $\text{K}$ or $\neg\text{K}$, $not$ is negation as failure. The set of all objective literals appears in an epistemic specification $\Pi$ is denoted by $Lit_\Pi$. Given a rule $r$ in the above form, let $head(r)$ denote its \emph{head} $\{l_0,...,l_k\}$, and $body(r)$ the \emph{body} $\{l_{k+1},...,l_{j},~\text{S}l_{j+1},...,~\text{S}l_{m},\;not\;l_{m+1},...,\;not\;l_{n}\}$. Furthermore, let $body^P(r)$ be the \emph{positive objective body} $\{l_{k+1},...,l_{g}\}$ and $body^N(r)$ \emph{negative objective body} $\{l_{m+1},...,l_n\}$ of $r$, and $body^S(r)$ the \emph{subjective body} $\{l_{j+1},...,l_{m}\}$. In addition, we use $body^K(r)$ to denote the set of objective literals in the body of $r$ which appears in term K, and $body^{-K}(r)$ to denote the set of objective literals in the body of $r$ which appears in term $\neg$K.

Epistemic specifications with variables are considered as shorthands for their ground instantiations. In the rest of this section, except special noted, we always consider the epistemic specification is grounded.

Let $W$ be a non-empty collection of sets of objective literals, and $l$ an objective literal.
\begin{itemize}
\item[-] {K}$l$ is satisfied with regard to $W$, denoted by $W\models$K$l$ , iff $\forall \omega\in W$: $l\in\omega$.
\item[-] $\neg${K}$l$ is satisfied with regard to $W$, denoted by $W\models\neg$K$l$ , iff $\exists\omega\in W$: $l\notin\omega$.
\end{itemize}

\begin{definition} Let $\Pi$ be an epistemic specification and $W$ be a non-empty collection of sets of objective literals in $\Pi$. $W$ is a world view of $\Pi$ iff $W$ is the collection of all answer sets of $\Pi^{W}$ denoted by $AN(\Pi^W)$, where $\Pi^{W}$ is an ASP program obtained from $\Pi$ by the following reduct laws:
\begin{itemize}
  \item[-] {RL1}: removing all rules containing subjective literals not satisfied by $W$;
  \item[-] {RL2}: removing any remaining subjective literals of the form $\neg${K}$l$;
  \item[-] {RL3}: replacing any remaining subjective literals of the form {K}$l$ by $l$.
\end{itemize}
\end{definition}

\ignore{\begin{example}
Given an epistemic specification $\Pi_1$:
\[l~:-~\mbox{K}l.\]
For $\emptyset$, because K $l$ is not satisfied wrt $\emptyset$, $\Pi_1^\emptyset=\emptyset$ and then $AN(\Pi_1^\emptyset)=\{\emptyset\}$ holds (by RL1). Hence,  $\emptyset$ is equal to $AN(\Pi_1^\emptyset)$. According to the definition of the world view, $\emptyset$ is a world view of $\Pi_1$. For $\{\{l\}\}$, because K $l$ is satisfied wrt $\{\{l\}\}$, $\Pi^{\{\{l\}\}}=\{l\leftarrow\;l\}$ holds (by RL3). Then, $AN(\Pi_1^{\{\{l\}\}})=\{\emptyset\}$ and $AN(\Pi_1^{\{\{l\}\}})\ne\{\{l\}\}$. Thus, $\{\{l\}\}$ is not a world view of $\Pi_1$.
\end{example}}

\begin{example}
Let an epistemic specification $\Pi_1$ consist of the following three rules:
\[p\;\mbox{or}\;q.~~~p\;:-\;\neg\mbox{K}\;q.~~~q\;:-\;\neg\mbox{K}\;p.\]
With regard to $\{\{p\}\}$, $\neg\mbox{K}\;q$ is satisfied while $\neg\mbox{K}\;p$ is not satisfied. Hence, $\Pi_1^{\{\{p\}\}}=\{p\;\mbox{or}\;q.\quad p\;:-.\}$ and then $AN(\Pi_1^{\{\{p\}\}})=\{\{p\}\}$. So $\{\{p\}\}$ is a world view of $\Pi_1$. Similarly, $\{\{q\}\}$ is also a world view of $\Pi_1$.
\end{example}

\subsection{Representation of Other Subjective Literals}
To handle other subjective literals using $ESmodels$, namely K $not~l$, $\neg$K $not~l$, M$l$, $\neg$M$l$, M $not~l$, and $\neg$M $not~l$, we can convert an epistemic specification $\Pi$ with arbitrary subjective literals in rules bodies into an epistemic specification $\Pi^{ES}$ such that $\Pi^{ES}$ has only subjective literals in the form K$l$ or $\neg$K$l$ by the following transformation procedure.
\begin{itemize}
  \item[1] For each objective literal $l$, add a rule $l^\prime:-~not~l$ to $\Pi^{ES}$ if there exist a subjective occurrence of $\neg$K $not~l$ or M$l$ or $\neg$M$l$ or K $not~l$ in $\Pi$, where $l^\prime$ is a new created objective literal corresponding to $l$.
  \item[2] {Add each rule of $\Pi$ to $\Pi^{ES}$ after performing the following operations on it.
  \begin{itemize}
    \item[-] Replace $\neg$K $not~l$ by $\neg$K$l^\prime$;
    \item[-] Replace M$l$ by $\neg$K$l^\prime$;
    \item[-] Replace $\neg$M$l$ by K$l^\prime$;
    \item[-] Replace K $not~l$ by K$l^\prime$;
    \item[-] Replace M $not~l$ by $\neg$K$l$;
    \item[-] Replace $\neg$M $not~l$ by K$l$.
  \end{itemize}}
\end{itemize}
Then, we define its \emph{world view} based semantics as follows.

\begin{definition}
For an epistemic specification $\Pi$ with arbitrary subjective literals, let $Lit$ be a set of objective literals appearing in $\Pi$, and $\Pi^{ES}$ its corresponding $ESmodels$ epistemic specification, a collection of sets of objective literals $W$ is a world view of $\Pi$ iff there exists a world view $W^\prime$ of $\Pi^{ES}$ such that $W=\{\omega\cap Lit|\omega\in W^\prime\}$.
\end{definition}

\begin{example} Given an epistemic specification $\Pi_2:\{p~:-~\neg\mbox{M} q.~~~q~:-~\neg\mbox{K} p.\}$
then we have $Lit_2=\{p, q\}$ and $\Pi_2^{ES}: \{p :- \mbox{K} l.~~~l :- not\;q.~~~q~:-~\neg\mbox{K} p.\}$.
$\Pi_2^{ES}$ has two world views $\{\{q\}\}$ and $\{\{p,l\}\}$, hence, $\Pi_2$ has two world views $\{\{q\}\}$ and $\{\{p\}\}$.
\end{example}

\begin{example} Given an epistemic specification $\Pi_3:\{ p~:-~not\;q,\mbox{M} q.~~~q~:-~not\;p,\mbox{M} q.\}$,
then we have $\Pi_3^{ES}:\{p :- not\;q,\neg\mbox{K} l.~~~l :- not\;q.~~~q :- not\;p,\neg\mbox{K} i.~~~i :- not\; q.\}$.
$\Pi_3^{ES}$ has two world views $\{\{i,l\}\}$ and $\{\{i,p,l\},\{q\}\}$, hence, $\Pi_3$ has two world views $\{\{\}\}$ and $\{\{p\},\{q\}\}$.
\end{example}

\subsection{Connection to Gelfond's New Epistemic Specification}

In the syntactic aspect of the epistemic specification defined in \cite{gelfond2011new}, it allows two more subjective literals of forms, K not $l$ and $\neg$K not $l$, in the rule's body. The modality M is defined to be expressed in terms of K by $\mbox{M}\;l=_{def}\neg\mbox{K}~not\;l$. Semantically, let $W$ be a non-empty collection of sets of objective literals, and $l$ an objective literal.
\begin{itemize}
\item[-] {K}$l$ is satisfied with regard to $W$, denoted by $W\models${K}$l$ , iff $\forall S\in W$: $l\in S$.
\item[-] $\neg${K}$l$ is satisfied with regard to $W$, denoted by $W\models\neg${K}$l$ , iff $\exists S\in W$: $l\notin S$.
\item[-] {K not} $l$ is satisfied with regard to $W$, denoted by $W\models\text{ K not } l$ iff for every $S\in W$, $l\notin S$, otherwise $S\models\neg\text{ K not } l$
\end{itemize}
The set $W$ is called a world view of $\Pi$ if $W$ is the collection of all answer sets of $\Pi^W$, where $\Pi^W$ is obtained by
\begin{itemize}
\item[-]removing all rules containing subjective literals not satisfied by $W$;
\item[-]removing any remaining subjective literals of the form $\neg${K}$l$ or $\neg$K$not~l$;
\item[-]replacing any remaining subjective literals of the form {K}$l$ by $l$ and any K$not~l$ by $not~l$.
\end{itemize}

Theorem\ref{theorem_connection} shows that $ESmodels$ can compute the world view of any Gelfond's new epistemic specification.
\begin{theorem}\label{theorem_connection}
For any Gelfond's new epistemic specification $\Pi$, let $Lit$ be a set of objective literals appearing in $\Pi$, a collection of sets of objective literals $W$ is a world view of $\Pi$ under Gelfond's new definition iff there exists a world view $W^\prime$ of $\Pi^{ES}$ such that $W=\{S\cap Lit|S\in W^\prime\}$.
\end{theorem}
\begin{proof} The main idea of this proof is as follows. Let $Lit^{ES}$ be objective literals appearing in $\Pi^{ES}$, \\
$\leftarrow$ direction: if there is a world view $W^\prime$ of $\Pi^{ES}$, then for any $\omega\in W^\prime$, $\omega$ is an answer set of $(\Pi^{ES})^{W^\prime}$. Let $W=\{S\cap Lit|S\in W^\prime\}$, then $\omega\cap Lit$ is an answer set of $\Pi^W$ under Gelfond's new definition (because the Gelfond-Lifschitz reduction of $\Pi^W$ wrt. $\omega\cap Lit$ just possibly has less facts $\{l:-.|l\in\omega-Lit~\text{and}~l~\text{does not appear in bodies of any rules}\}$ than the Gelfond-Lifschitz reduction of $(\Pi^{ES})^{W^\prime}$ wrt. $\omega$).
\\
$\rightarrow$ direction: if $W$ is a world view of $\Pi$, then we create $W^\prime$ as follows: for each $\omega\in W$, we have $\omega^\prime=\omega\cup\{l\in Lit^{ES}-Lit|l:-not~l^\prime\in\Pi^{ES}, l^\prime\notin\omega\}$ in $W^\prime$. Then, $\omega^\prime$ is an answer set of $(\Pi^{ES})^{W^\prime}$ (because the Gelfond-Lifschitz reduction of $(\Pi^{ES})^{W^\prime}$ wrt. $\omega^\prime$ just possibly has more facts $\{l:-.|l\in\omega-Lit~\text{and}~l~\text{does not appear in bodies of any rules}\}$ than the Gelfond-Lifschitz reduction of $\Pi^W$ wrt. $\omega$).
\end{proof}
\begin{example} Given an epistemic specification $\Pi_4: \{p :- \mbox{M} p.\}$, under Gelfond's definition $\Pi_4$ has two world views $\{\{\}\}$ and $\{\{p\}\}$. By the transformation defined in last subsection, we have $\Pi_4^{ES}:\{p :- \neg\mbox{K} l.~~~:- not\;p.\}$, and $ESmodels$ can find $\Pi_4^{ES}$ 's two world views: $\{\{l\}\}$ and $\{\{p\}\}$, that is, $\Pi_4$ also has two world views $\{\{\}\}$ and $\{\{p\}\}$ by $ESmodels$.
\end{example}

\ignore{\subsection{Elimination of Seft-recursion Through \emph{M}}
From example 5, we find that the recursion through M. ??$How is M defined in Gelfond's slides$??

{\noindent}{\underline {Example 5}}. $\Pi_5:$
\[ p~:-~Mp.\]
then we have $\Pi_5^{ES}:$
\[p :- \neg\mbox{K} l.~~~l :- not\;p.\]
$\Pi_5^{ES}$ has two world views $\{\{p\}\}$ and $\{\{l\}\}$, hence, $\Pi_4$ has two world views $\{\{\}\}$ and $\{\{p\},\{q\}\}$.
} 

\section{Computing World Views in $ESmodels$}
A generate-test algorithm forms a basis of computing world views in $ESmodels$. Now, we are taking two preliminary steps to optimize the algorithm.
\subsection{Basic Algorithm}
Let $\Pi$ be an epistemic specification, $EL(\Pi)$ be a set of objective literals such that $l\in EL(\Pi)$ iff K$l$ or -K$l$ occurring in $\Pi$.  Then, we call a pair $(S,S^\prime)$ an \emph{assignment} of $EL(\Pi)$ iff
\[S\cup S^\prime=EL(\Pi)\text{ and } S\cap S^\prime=\emptyset\]
Then, we define an answer set program $\Pi^{(S, S^\prime)}$ obtained by:
\begin{itemize}
\item[-]removing from $\Pi$ all rules containing subjective literals K$l$ such that $l\in S^\prime$, or subjective literal $\neg$K$l$ such that $l\in S$,
\item[-]removing from the rest rules in $\Pi$ all other occurrences of subjective literals of the form $\neg$K$l$,
\item[-]replacing remaining occurrences of literals of the form K$l$ by $l$.
\end{itemize}
\begin{theorem}\label{basic_alg} Given an epistemic specification $\Pi$ and a collection $W$ of sets of objective literals. $W$ is a world view of $\Pi$ if an assignment $(S,S^\prime)$ of $EL(\Pi)$ exists such that
\begin{itemize}
\item[-]$W$ is the collection of all answer sets of $\Pi^{(S, S^\prime)}$,
\item[-]$W$ satisfies the assignment, that is, $S\cap\big(\bigcap_{A\in W}\big)==S$ and $S^\prime\cap\big(\bigcap_{A\in W}\big)==\emptyset$.
\end{itemize}
\end{theorem}
\begin{proof} If both $S\cap\big(\bigcap_{A\in W}\big)==S$ and $S^\prime\cap\big(\bigcap_{A\in W}\big)==\emptyset$ are satisfied, we have $\Pi^{(S, S^\prime)}=\Pi^W$. Hence, if $W$ is the collection of all answer sets of $\Pi^{(S, S^\prime)}$ then $W$ is the collection of all answer sets of $\Pi^W$, that is, $W$ is a world view of $\Pi$.
\end{proof}

By Theorem \ref{basic_alg}, an immediate method of computing the world views of an epistemic specification includes three main stages: generating a possible assignment, reducing the epistemic specification into an answer set program, and testing if the collection of the answer sets of the answer set program satisfies the assignment. At a high level of abstraction, the method can be implemented as showed in the following algorithm.

\begin{algorithm}\label{basic_algorithm}
\renewcommand{\algorithmicrequire}{\textbf{Input:}}
\renewcommand\algorithmicensure {\textbf{Output:} }
\caption{ ESMODELS.}
\begin{algorithmic}[1]
\REQUIRE ~~\\
$\Pi$: An epistemic specification;\\

\ENSURE ~~\\ 
All world views of $\Pi$;\\

\FOR{ every possible assignment of $EL(\Pi)$ $(S, S^\prime)$ of $\Pi$}
\STATE $\Pi^{'} = \Pi^{( S, S^\prime)}$          \COMMENT{reduces $\Pi$ to an answer set program $\Pi^{'}$ by $(S, S^\prime)$}
\STATE $W =$ computerASs($\Pi^{'}$)              \COMMENT{computes all answer sets of $\Pi^{'}$}
\IF {$S\cap\big(\bigcap_{A\in W}\big)==S$ and $S^\prime\cap\big(\bigcap_{A\in W}\big)==\emptyset$}
\STATE output $W$
\ENDIF
\ENDFOR
\end{algorithmic}
\end{algorithm}
ESMODELS firstly gets all subjective literals $EL(\Pi)$ and generates all possible assignments of $EL(\Pi)$.
For each assignment$(S, S^\prime)$, the algorithm reduces $\Pi$ to an answer set program $\Pi^\prime$, i.e., $\Pi^\prime=\Pi^{(S, S^\prime)}$.
Next, it calls exiting ASP solver like Smodels, Clasp to compute all answer sets $W$ of $\Pi^\prime$. Finally, it verifies the $W$. $W$ is a world view of $\Pi$, if $W$ satisfies $S\cap\big(\bigcap_{A\in W}\big)==S$ and $S^\prime\cap\big(\bigcap_{A\in W}\big)==\emptyset$. ESMODELS stops, when all possible assignments are tested.
\subsection{Optimization Approaches}
\subsubsection{Reducing Subjective Literals}
However, ESMODELS has a high computational cost, especially with a large number of subjective literals. Therefore, we introduce a new preprocessing function to reduce reduce $EL(\Pi)$ before generating all possible assignments of $EL(\Pi)$. We first give several propositions.

Let $\Pi$ be an epistemic specification and a pair $(S,S^\prime)$ of objective literals of $\Pi$, $T_\Pi$ be an \emph{lower bound} operator on $(S,S^\prime)$ defined as follows:
\begin{center}
$T_\Pi(S,S^\prime)=\big(\{head(r)||head(r)|=1,body^+(r)\subseteq S, body^-(r)\subseteq S^\prime\},$\\
$\{l|\neg\exists r\in\Pi(l\in head(r)), \text{ or }\forall r\in\Pi, l\in head(r)\Rightarrow(body^+(r)\cap S^\prime\neq\emptyset\text{ or }body^-(r)\cap S\neq\emptyset)\}\big)$
\end{center}
where $body^+(r)=body^P(r)\cup body^K(r)$, $body^-(r)=body^N(r)\cup body^{-K}(r)$. Intuitively, $T_\Pi(S,S^\prime)$ computes the objective literals that must be true and that not true with regard to $S$ and $S^\prime$ which are sets of literals known true and known not true respectively. Clearly, we can use this operation to reduce the searching space of subjective literals. This idea is guaranteed by the following definitions and propositions.
\begin{definition}\label{Partail Model}
A pair $(S,S^\prime)$ of sets of objective literals is a partial model of an epistemic specification $\Pi$ if, for any world view $W$ of $\Pi$, $S\cap\big(\bigcap_{A\in W}\big)==S$ and $S^\prime\cap\big(\bigcap_{A\in W}\big)==\emptyset$.
\end{definition}
\begin{theorem}
$T_\Pi(S, S^\prime)$ is a partial model if $(S, S^\prime)$ is a partial model of an epistemic specification $\Pi$, .
\end{theorem}
\begin{proof}Let $(A, B)|_1$ to denote $A$ of a pair $(A, B)$, and $(A, B)|_2$ to denote $B$. The main idea of this proof is as follows. For any world view $W$ of $\Pi$, $S\cap\big(\bigcap_{A\in W}\big)==S$ and $S^\prime\cap\big(\bigcap_{A\in W}\big)==\emptyset$, by the definition of $T_\Pi$, the Gelfond-Lifschitz reduction of $\Pi^W$ wrt. any $\omega\in W$ must have ${l:-|l\in T_\Pi(S, S^\prime)|_1}$ and must not have any rule with head in $T_\Pi(S, S^\prime)|_2$, hence, we have $T_\Pi(S, S^\prime)|_1\cap\big(\bigcap_{A\in W}\big)==T_\Pi(S, S^\prime)|_1$ and $T_\Pi(S, S^\prime)|_2\cap\big(\bigcap_{A\in W}\big)==\emptyset$.
\end{proof}
\begin{corollary}
Let, $T_\Pi^i(S,S^\prime)=T_\Pi(T_\Pi^{i-1}(S,S^\prime))$, then $T_\Pi^k(\emptyset,\emptyset)$ is a partial model of $\Pi$.
\end{corollary}
\begin{proof}
Because $(\emptyset, \emptyset)$ is a partial model, $T_\Pi(\emptyset,\emptyset)$ is a partial model, and so on, $T_\Pi^2(\emptyset,\emptyset)$ ... $T_\Pi^k(\emptyset,\emptyset)$ are partial models of $\Pi$
\end{proof}
An epistemic specification rule $r$ is \emph{defeated} by $(S, S^\prime)$ if $body^+(r)\cap S^\prime\neq\emptyset$ or $body^-(r)\cap S^\prime\neq\emptyset$. Let $(S, S^\prime)$ be a partial model of an epistemic specification $\Pi$, $\Pi|_{(S, S^\prime)}$ is obtained by
\begin{itemize}
\item[-]removing from $\Pi$ all rules defeated by $(S, S^\prime)$,
\item[-]removing from the rest rules in $\Pi$ all other occurrences of literals of the form not $l$ or $\neg$K$l$ such that $l\in S^\prime$,
\item[-]removing remaining occurrences of literals of the form $l$ or K$l$ such that $l\in S$.
\item[-]adding $l\leftarrow.$ if $l\in S$
\item[-]adding $\leftarrow l.$ if $l\in S^\prime$
\end{itemize}

\begin{theorem}
If $(S, S^\prime)$ is a partial model of an epistemic specification $\Pi$, $\Pi|_{(S, S^\prime)}$ and $\Pi$ have the same world views.
\end{theorem}
\begin{proof}
The main idea in this proof is as follows. For any world view $W$ of $\Pi$, if $S\cap\big(\bigcap_{A\in W}\big)==S$ and $S^\prime\cap\big(\bigcap_{A\in W}\big)==\emptyset$, then $\Pi^W$ and $(\Pi|_{(S, S^\prime)})^W$ have the same answer sets. And, for any world view $W$ of $\Pi|_{(S, S^\prime)}$, we have that $W$ is a world view of $\Pi$.
\end{proof}

By theorem 3 and 4, we can design PreProcess showed in algorithm 2. Firstly, it sets the pair ($S,S^{'}$) as $(\emptyset,\emptyset)$. Then it expands the partial model of $\Pi$ and reducts the $\Pi^{'}$ according to ($S,S^{'}$). Next, we updates the partial model by the new program. Finally, it compares the new partial model with the previous one. If the partial model is stable, it stops and returns $\Pi^{'}$; Otherwise, it repeats this procedure.
\begin{algorithm}
\renewcommand{\algorithmicrequire}{\textbf{Input:}}
\renewcommand\algorithmicensure {\textbf{Output:} }
\caption{ PreProcess.}
\begin{algorithmic}[1]
\REQUIRE ~~\\
$\Pi$: An epistemic specification;\\

\ENSURE ~~\\ 
$\Pi^{'}$: A reduction of $\Pi$;\\

\STATE $(S, S^\prime)=(\emptyset,\emptyset)$,
\REPEAT
\STATE $(S,S^\prime)=T_{\Pi^\prime}(S, S^\prime)$
\STATE $\Pi^\prime =\Pi^\prime|_{(S,S^\prime)}$
\UNTIL {$S,S^\prime$ are fixed}
\STATE {return $\Pi^\prime$}
\end{algorithmic}
\end{algorithm}

Obviously, PreProcess and partial model are very helpful for reducing search space. We thus provide an EFFICIENT ESMODELS as follows:
\begin{algorithm}
\renewcommand{\algorithmicrequire}{\textbf{Input:}}
\renewcommand\algorithmicensure {\textbf{Output:} }
\caption{ EFFICIENT ESMODELS.}
\begin{algorithmic}[1]
\REQUIRE ~~\\
$\Pi$: An epistemic specification;\\

\ENSURE ~~\\ 
All world views of $\Pi$;\\
\STATE $\Pi^\prime$=PreProcess($\Pi$)
\FOR{ every possible assignment of $EL(\Pi^\prime)$}
\STATE $\Pi^\prime = {\Pi^\prime}^{(S, S^\prime)}$
\STATE $\Pi^\prime$=PreProcess($\Pi^\prime$)
\STATE $W =$ computerASs($\Pi^\prime$)
\IF {$S\cap\big(\bigcap_{A\in W}\big)==S$ and $S^\prime\cap\big(\bigcap_{A\in W}\big)==\emptyset$}
\STATE output $W$
\ENDIF
\ENDFOR
\end{algorithmic}
\end{algorithm}

\subsubsection{Using Multicore Technology}
In $ESmodels$, another way of improving efficiency is the use of multicore technology. Based on Algorithm 3, by parallel generation of possible assignments and parallel calling of ASP solver, the efficiency of $ESmodels$ can be improved greatly. 

\section{Applications}
\ignore{\subsection{Commonsense Reasoning}}
\subsection{Conformant Planning}
Consider the planning problem with multiple possible initial states, what makes it become much harder is to find a so called \emph{secure} plan that enforces the goal from any initial state. \cite{eiter2003logic} gives three security conditions to check whether a plan is secure:
\ee{
\item the actions of the plan are executable in the respective stages of the execution;
\item at any stage, executing the respective actions of the plan always leads to some legal successor state; and
\item the goal is true in every possible state reached if all steps of the plan are successfully executed.
}
Here, we consider a track of effects of executing an action sequence as a belief set, thus can intuitively encode those security conditions in epistemic specification constraints. We use $nonexecutable$ to denote the actions are not executable, $inconsistent$ to denote that a state is illegal, $success$ to sign a state satisfies the goal, and $goal(m)$ to denote the state reached after a given steps number $m$ satisfies the goal, and $o(A, T)$ to denote an action $A$ happens in the step $T$:
\ie{
\item[-] for security condition 1:$~~~\leftarrow M~nonexecutable.$
\item[-] for security condition 2:$~~~\leftarrow M~inconsistent.$
\item[-] for security condition 3:$~~~success\leftarrow goal(m).$ and $\leftarrow \neg K~success.$
}
\ignore{Naturally, the constraint \eqref{cc3} says a plan should be executable in any stage under any execution track starting from any possible initial state, and rule \eqref{cc4} expresses that any action in a plan never causes inconsistency under any execution track, that is, a plan always leads to legal successor state, and \eqref{cc0} and \eqref{cc1} together require a plan should always take the system to a goal state.}
Moreover, to guarantee the above security testing is put on tracks caused by the same action sequence, we write a new constraint.
\begin{equation}\label{cc2}\leftarrow \neg K o(A, T), o(A, T).\end{equation}
Intuitively, rule \eqref{cc2} says that \emph{if one action $A$ happened in stage $T$ of one track, it happened in stage $T$ of all tracks}. Thus, we can easily get a {\bf Conformant Planning Module} consisting of the above five constraints and the following action generation rules:
   \ie{
     \item Set a planning horizon $m$: $\#const~x=m.$ $step(0..x).$
     \item Generating one action for each step: $1\{o(A, T):action(A)\}1\leftarrow step(T), T<m.$
   }
Combine the conformant planning module with a planning domain (including action axioms e.g., inertial law) encoded in an answer set program, the result epistemic specification represents a conformant planning problem, and its world view(s) corresponds to the secure plan(s) of the problem. Here, we use a case provided in \cite{PalaciosG06} to demonstrate the conformant planning approach using epistemic specification.
Given a conformant planning problem $P$ with an initial state $I={p\vee q}$ (i.e., nothing else is known; there is no CWA), and action $a$ and $b$ with effects $a$ causes $q$ if $r$, $a$ causes $\neg s$ if $r$, and $b$ causes $s$ if $q$, the planning goal is ${q, s}$.
Then, we describe the planning domain as follows\ignore{\footnote{Interesting readers can find the complete epistemic specification at\\ http://cse.seu.edu.cn/PersonalPage/seu\_zzz/esmodels/cpes\_examples/p3.txt}}.
\ie{
\item Signatures:$~~action(a).~~action(b).$\\
$~~~~~~~~~~~~~~fluent(in, p).~~fluent(in, q).~~fluent(in, r).~~fluent(in, s).$
\item Causal Laws:$~~h(pos(q), T+1):- o(a, T),h(pos(p),T),step(T).$\\
$~~~~~~~~~~~~~~h(neg(s), T+1):- o(a, T),h(pos(r),T),step(T).$\\
$~~~~~~~~~~~~~~h(pos(s), T+1):- o(b, T),h(pos(q),T),step(T).$
\item Inertial~~Laws:\\$h(pos(X), T+1):- fluent(in, X),h(pos(X),T),step(T),not~h(neg(X),T+1).$\\
         $h(neg(X), T+1):- fluent(in, X),h(neg(X),T),step(T),not~h(pos(X),T+1).$
\item Initial:$~~1\{h(pos(p),0), h(pos(q),0)\}2.$\\
$~~~~~~~~~~~~~~1\{h(pos(F), 0), h(neg(F), 0)\}1 :- fluent(in, F).$
 \item Goal:$~~goal(T):- h(pos(q), T),h(pos(s), T),step(T).$
}
When we set $m=2$, $ESmodels$ can find the unique world view including twelve literal sets, and each of them includes $o(a, 0)$ and $o(b, 1)$ that means the program has a conformant plan $a~b$.
\subsection{Constraints Satisfaction}
In some situations, constraints on the variable are with epistemic features, that is, a variable's value is not only affected by the values of other variables, but also determined by all possible values of other variables. Here, we demonstrate the use of $ESmodels$ in solving such constraint satisfaction problems using a \emph{dinner} problem:Jim, Bones, Checkov, Mike, Jack, Uhura, and Scotty, and Tommy received a dinner invitation, and the constraints on their decisions and the constraints description in epistemic specification rules are as follows:
\ie{
\item if Checkov may not participate, then Jim will participate: $jim :- ~not~checkove.$
\item if Jim may not participate, then bones will participate: $bones :- ~not~jim.$
\item if only one of Jack and Mike will participate: $jack:-~not~mike.~~~mike:-~not~jack.$
\item if Jack must participate, then Uhura will participate:
$uhura:-\text{K}jack.$
\item if Uhura may not participate, then Scotty will participate:
$scotty :- ~not~uhura.$
\item if Scotty must participate, then Tommy will participate:
$tommy:-\text{K}scotty.$
\item Checkov will participate.
$checkov.$
}
$ESmodels$ can find the unique world view$\{\{checkov, tommy, scotty, jim, mike\}\\\{checkov, tommy, scotty, jim, jack\}\}$ that means Jim, Checkov, Scotty, and tommy must participate, Bones and Uhura must not participate, Jack and Mike may or may not participate.

\section{Conclusion}
$ESmodels$ is an epistemic specification solver designed and implemented as an experiment platform to investigate the semantics, language, related reasoning algorithms, and possible applications of epistemic specifications. A significant feature of this solver is that its language is more compact than that defined in literatures, but capable of representing many subjective literals via a group of transformation rules. Besides, this solver can compute world views under Gelfond's new definition, while that presented by Zhang in \cite{zhang2007epistemic} and Watson in \cite{watson1994} are based on the early definition of epistemic specifications. In addition, we find the compact encoding of conformant planning problems and constraint satisfaction problems in the epistemic specification language, which primarily shows $ESmodels$'s potential in applications\footnote{In the early related work, Gelfond investigated the value of epistemic specifications in formalizing commonsense reasoning}.

The work presented here is primary. Now, we are designing and exploring more efficient algorithm for $ESmodels$ and evaluate it using those benchmarks in the conformant planning field. 

\section*{Acknowledgment}
We acknowledge the support from Project 60803061 and 61272378 by National Natural Science Foundation of China, and Project BK2008293 by Natural Science Foundation of Jiangsu.

\bibliographystyle{acmtrans}
\bibliography{iclp14_corr}

\begin{thebibliography}{}

\bibitem[\protect\citeauthoryear{Eiter, Faber, Leone, Pfeifer, and
  Polleres}{Eiter et~al\mbox{.}}{2003}]{eiter2003logic}
{\sc Eiter, T.}, {\sc Faber, W.}, {\sc Leone, N.}, {\sc Pfeifer, G.}, {\sc and}
  {\sc Polleres, A.} 2003.
\newblock A logic programming approach to knowledge-state planning, ii: The
  dlvk system.
\newblock {\em Artificial Intelligence\/}~{\em 144,\/}~1, 157--211.

\bibitem[\protect\citeauthoryear{Faber and Woltran}{Faber and
  Woltran}{2009}]{faber2009manifold}
{\sc Faber, W.} {\sc and} {\sc Woltran, S.} 2009.
\newblock Manifold answer-set programs for meta-reasoning.
\newblock In {\em Logic Programming and Nonmonotonic Reasoning}. Springer,
  115--128.

\bibitem[\protect\citeauthoryear{Faber and Woltran}{Faber and
  Woltran}{2011}]{faber2011manifold}
{\sc Faber, W.} {\sc and} {\sc Woltran, S.} 2011.
\newblock Manifold answer-set programs and their applications.
\newblock In {\em Logic programming, knowledge representation, and nonmonotonic
  reasoning}. Springer, 44--63.

\bibitem[\protect\citeauthoryear{Gelfond}{Gelfond}{1991}]{491768}
{\sc Gelfond, M.} 1991.
\newblock {Strong Introspection}.
\newblock In {\em National Conference on Artificial Intelligence}. 386--391.

\bibitem[\protect\citeauthoryear{Gelfond}{Gelfond}{1994}]{gelfond1994logic}
{\sc Gelfond, M.} 1994.
\newblock Logic programming and reasoning with incomplete information.
\newblock {\em Annals of mathematics and artificial intelligence\/}~{\em
  12,\/}~1-2, 89--116.

\bibitem[\protect\citeauthoryear{Gelfond}{Gelfond}{2011}]{gelfond2011new}
{\sc Gelfond, M.} 2011.
\newblock New semantics for epistemic specifications.
\newblock In {\em Logic Programming and Nonmonotonic Reasoning}. Springer,
  260--265.

\bibitem[\protect\citeauthoryear{Gelfond and Przymusinska}{Gelfond and
  Przymusinska}{1991}]{GelfondP91}
{\sc Gelfond, M.} {\sc and} {\sc Przymusinska, H.} 1991.
\newblock Definitions in epistemic specifications.
\newblock In {\em LPNMR} (2002-01-03). 245--259.

\bibitem[\protect\citeauthoryear{Gelfond and Przymusinska}{Gelfond and
  Przymusinska}{1993}]{gelfond1993reasoning}
{\sc Gelfond, M.} {\sc and} {\sc Przymusinska, H.} 1993.
\newblock Reasoning on open domains.
\newblock In {\em LPNMR}. Vol. 1993. 397--413.

\bibitem[\protect\citeauthoryear{Palacios and Geffner}{Palacios and
  Geffner}{2006}]{PalaciosG06}
{\sc Palacios, H.} {\sc and} {\sc Geffner, H.} 2006.
\newblock Compiling uncertainty away: Solving conformant planning problems
  using a classical planner (sometimes).
\newblock In {\em AAAI}. AAAI Press, 900--905.

\bibitem[\protect\citeauthoryear{Truszczy{\'n}ski}{Truszczy{\'n}ski}{2011}]{tr%
uszczynski2011revisiting}
{\sc Truszczy{\'n}ski, M.} 2011.
\newblock Revisiting epistemic specifications.
\newblock In {\em Logic programming, knowledge representation, and nonmonotonic
  reasoning}. Springer, 315--333.

\bibitem[\protect\citeauthoryear{Wang and Zhang}{Wang and
  Zhang}{2005}]{wang2005nested}
{\sc Wang, K.} {\sc and} {\sc Zhang, Y.} 2005.
\newblock Nested epistemic logic programs.
\newblock In {\em Logic Programming and Nonmonotonic Reasoning}. Springer,
  279--290.

\bibitem[\protect\citeauthoryear{Watson}{Watson}{1994}]{watson1994}
{\sc Watson, R.} 1994.
\newblock An inference engine for epistemic specifications.
\newblock {\em 1994.M.S. Thesis, Department of Computer Science, University of
  Texas at El Paso.\/}.

\bibitem[\protect\citeauthoryear{Watson}{Watson}{2000}]{watson2000splitting}
{\sc Watson, R.} 2000.
\newblock A splitting set theorem for epistemic specifications.
\newblock {\em In Proceedings of the 8th International Workshop on
  Non-MonotonicReasoning (NMR-2000)\/}.

\bibitem[\protect\citeauthoryear{Zhang}{Zhang}{2003}]{zhang2003minimal}
{\sc Zhang, Y.} 2003.
\newblock Minimal change and maximal coherence for epistemic logic program
  updates.
\newblock In {\em IJCAI}. 112--120.

\bibitem[\protect\citeauthoryear{Zhang}{Zhang}{2006}]{zhang2006computational}
{\sc Zhang, Y.} 2006.
\newblock Computational properties of epistemic logic programs.
\newblock In {\em KR}. 308--317.

\bibitem[\protect\citeauthoryear{Zhang}{Zhang}{2007}]{zhang2007epistemic}
{\sc Zhang, Y.} 2007.
\newblock Epistemic reasoning in logic programs.
\newblock In {\em IJCAI}. 647--653.

\end{thebibliography}

\label{lastpage}
\end{document}